# Two-Manifold Problems with Applications to Nonlinear System Identification

**Byron Boots** BEB@CS.CMU.EDU
**Geoffrey J. Gordon** GGORDON@CS.CMU.EDU
Machine Learning Department, Carnegie Mellon University, Pittsburgh, PA 15213

## Abstract

Recently, there has been much interest in spectral approaches to learning manifolds—so-called kernel eigenmap methods. These methods have had some successes, but their applicability is limited because they are not robust to noise. To address this limitation, we look at *two-manifold problems*, in which we simultaneously reconstruct two related manifolds, each representing a different view of the same data. By solving these interconnected learning problems together, two-manifold algorithms are able to succeed where a non-integrated approach would fail: each view allows us to suppress noise in the other, reducing bias. We propose a class of algorithms for two-manifold problems, based on spectral decomposition of cross-covariance operators in Hilbert space, and discuss when two-manifold problems are useful. Finally, we demonstrate that solving a two-manifold problem can aid in learning a nonlinear dynamical system from limited data.

## 1. Introduction

Manifold learning algorithms are nonlinear methods for embedding a set of data points into a low-dimensional space while preserving local geometry. Recently, there has been a great deal of interest in spectral approaches to learning manifolds. These *kernel eigenmap* methods include Isomap (Tenenbaum et al., 2000), Locally Linear Embedding (LLE) (Roweis & Saul, 2000), Laplacian Eigenmaps (LE) (Belkin & Niyogi, 2002), Maximum Variance Unfolding (MVU) (Weinberger et al., 2004), and Maximum Entropy Unfolding (MEU) (Lawrence, 2011). Despite the popularity of kernel eigenmap methods, they are limited in one important respect: they generally only perform well when there is *little or no noise*. Several authors have attacked the problem of learning manifolds in the presence of noise using methods like neighborhood smoothing (Chen et al., 2008) and robust principal components analysis (Zhan & Yin, 2009; 2011), with some success when noise is limited. Unfortunately, the problem is fundamentally ill posed without some sort of side information about the true underlying signal: by design, manifold methods will recover extra latent dimensions which "explain" the noise.

We take a different approach to the problem of learning manifolds from noisy observations. We assume access to *instrumental variables*, which are correlated with the true latent variables, but uncorrelated with the noise in observations. Such instrumental variables can be used to separate signal from noise, as described in Section 3. Instrumental variables have been used to allow consistent estimation of model parameters in many statistical learning problems, including linear regression (Pearl, 2000), principal component analysis (PCA) (Jolliffe, 2002), and temporal difference learning (Bradtke & Barto, 1996). Here we extend the scope of this technique to manifold learning. We will pay particular attention to the **two-manifold problem**, in which two sets of observations each serve as instruments for the other. We propose algorithms for two-manifold problems based on spectral decompositions related to *cross-covariance operators*; and, we show that the instrumental variable idea suppresses noise in practice.

Finally we look at a detailed example of how two-manifold methods can help solve difficult machine learning problems. *Subspace identification* approaches to learning nonlinear dynamical systems depend critically on instrumental variables and the spectral decomposition of (potentially infinite-dimensional) covariance operators (Hsu et al., 2009; Boots et al., 2010; Song et al., 2010). Two-manifold problems are a natural fit: by relating the spectral decomposition to our two-manifold method, subspace identification techniques can be forced to identify a *manifold* state space, and consequently, to learn a dynamical system that is both accurate and interpretable, outperforming the current state of the art.

## 2. Preliminaries
### 2.1. Kernel PCA
Kernel PCA (Schölkopf et al., 1998) generalizes PCA to high- or infinite-dimensional input data, represented implicitly using a reproducing-kernel Hilbert space (RKHS). If the kernel $K(\mathbf{x}, \mathbf{x}')$ is sufficiently expressive, kernel PCA can find structure that regular PCA misses. Conceptually, if we write the "feature function" $\phi(\mathbf{x}) = K(\mathbf{x}, \cdot)$, and define an infinitely-tall "matrix" $\mathbf{\Phi}$ with columns $\phi(\mathbf{x}_i)$, our goal is to recover the eigenvalues and eigenvectors of the centered *covariance operator* $\hat{\mathbf{\Sigma}}_{XX} = \frac{1}{n}\mathbf{\Phi}\mathbf{H}\mathbf{\Phi}^\mathsf{T}$. Here $\mathbf{H}$ is the centering matrix $\mathbf{H} = \mathbf{I}_n - \frac{1}{n}\mathbf{1}\mathbf{1}^\mathsf{T}$. For efficient computation, we work with the *Gram matrix* $\mathbf{G} = \frac{1}{n}\mathbf{\Phi}^\mathsf{T}\mathbf{\Phi}$ instead of the large or infinite $\hat{\mathbf{\Sigma}}_{XX}$. The centered Gram matrix $\mathbf{C} = \mathbf{H}\mathbf{G}\mathbf{H}$ has the same nonzero eigenvalues as $\hat{\mathbf{\Sigma}}_{XX}$; the eigenvectors of $\hat{\mathbf{\Sigma}}_{XX}$ are $\mathbf{\Phi}\mathbf{H}\mathbf{v}_i\lambda_i^{-1/2}$, where $(\lambda_i, \mathbf{v}_i)$ are the eigenpairs of $\mathbf{C}$ (Schölkopf et al., 1998).

### 2.2. Manifold Learning
Kernel eigenmap methods seek a nonlinear function that maps a high-dimensional set of data points to a lower-dimensional space while preserving the manifold on which the data lies. The main insight behind these methods is that large distances in input space are often meaningless due to the large-scale curvature of the manifold; so, ignoring these distances can lead to a significant improvement in dimensionality reduction by "unfolding" the manifold.

Interestingly, these algorithms can be viewed as special cases of kernel PCA where the Gram matrix $\mathbf{G}$ is constructed over the finite domain of the training data in a particular way (Ham et al., 2003). For example, in Laplacian Eigenmaps (LE), we first compute an adjacency matrix $\mathbf{W}$ by nearest neighbors: $w_{i,j}$ is nonzero if point $i$ is one of the nearest neighbors of point $j$, or vice versa. We can either set non-zero weights to 1, or compute them with a kernel such as a Gaussian RBF. Next we let $\mathbf{S}_{i,i} = \sum_j w_{i,j}$, and set $\mathbf{L} = (\mathbf{S} - \mathbf{W})$. Finally, we eigendecompose $\mathbf{L} = \mathbf{V}\mathbf{\Lambda}\mathbf{V}^\mathsf{T}$ and set the LE embedding to be $\mathbf{E} = \mathbf{V}_{2:k+1}$, the $k$ smallest eigenvectors of $\mathbf{L}$ excluding the vector corresponding to the 0 eigenvalue. (Optionally we can scale according to eigenvalues, $\mathbf{E} = \mathbf{V}_{2:k+1}\mathbf{\Lambda}_{2:k+1}^{-1/2}$.) To relate LE to kernel PCA, Ham et al. (2003) showed that one can build a Gram matrix from $\mathbf{L}$, $\mathbf{G} = \mathbf{L}^\dagger$; the LE embedding is given by the top $k$ eigenvectors (and optionally eigenvalues) of $\mathbf{G}$.

In general, many manifold learning methods (including all the kernel eigenmap methods mentioned above) can be viewed as constructing a matrix $\mathbf{E}$ of embedding coordinates. From any such method, we can extract an equivalent Gram matrix $\mathbf{G} = \mathbf{E}\mathbf{E}^\mathsf{T}$. So, for the rest of the paper, we view a manifold learner simply as a black box which accepts data and produces a Gram matrix that encodes the learned manifold structure. This view greatly simplifies the description of our two-manifold algorithms below.

## 3. Bias and Instrumental Variables

Kernel eigenmap methods are very good at dimensionality reduction when the original data points sample a high-dimensional manifold relatively densely, and when the noise in each sample is small compared to the local curvature of the manifold. In practice, however, observations are frequently noisy, and manifold-learning algorithms applied to these datasets usually produce *biased* embeddings. See Figures 1–2, the "noisy swiss rolls," for an example.

Our goal is therefore to design a more noise-resistant algorithm for the two-manifold problem. We begin by examining PCA, a linear special case of manifold learning, and studying why it produces biased embeddings in the presence of noise. We next show how to overcome this problem in the linear case, and then use these same ideas to generalize kernel PCA, a nonlinear algorithm. Finally, in Sec. 4, we extend these ideas to fully general kernel eigenmap methods.

### 3.1. Bias in Finite-Dimensional Linear Models
Suppose that $\mathbf{x}_i$ is a noisy view of some underlying low-dimensional latent variable $\mathbf{z}_i$: $\mathbf{x}_i = \mathbf{M}\mathbf{z}_i + \boldsymbol{\epsilon}_i$ for a linear transformation $\mathbf{M}$ and i.i.d. zero-mean noise term $\boldsymbol{\epsilon}_i$.[1] Without loss of generality, we assume that $\mathbf{x}_i$ and $\mathbf{z}_i$ are centered, and that $\mathrm{Cov}[\mathbf{z}_i]$ and $\mathbf{M}$ both have full column rank. In this case, PCA on $\mathbf{X}$ will generally *fail* to recover $\mathbf{Z}$: the expectation of $\hat{\mathbf{\Sigma}}_{XX} = \frac{1}{n}\mathbf{X}\mathbf{X}^\mathsf{T}$ is $\mathbf{M}\,\mathrm{Cov}[\mathbf{z}_i]\,\mathbf{M}^\mathsf{T} + \mathrm{Cov}[\boldsymbol{\epsilon}_i]$, while we need $\mathbf{M}\,\mathrm{Cov}[\mathbf{z}_i]\,\mathbf{M}^\mathsf{T}$ to be able to recover a transformation of $\mathbf{M}$ or $\mathbf{Z}$. The unwanted term $\mathrm{Cov}[\boldsymbol{\epsilon}_i]$ will, in general, affect all eigenvalues and eigenvectors of $\hat{\mathbf{\Sigma}}_{XX}$, causing us to recover a *biased* answer even in the limit of infinite data.

#### 3.1.1. Instrumental Variables
We can fix this problem for linear embeddings: instead of plain PCA, we can use what might be called

---

[1] Note that each $\boldsymbol{\epsilon}_i$ is a vector, and we make no assumption that its coordinates are independent from one another (i.e., we do not assume that $\mathrm{Cov}[\boldsymbol{\epsilon}_i]$ is spherical).

*two-subspace* PCA. This method finds a statistically consistent solution through the use of an *instrumental variable* (Pearl, 2000; Jolliffe, 2002), an observation $\mathbf{y}_i$ that is correlated with the true latent variables, but uncorrelated with the noise in $\mathbf{x}_i$. Importantly, picking an instrumental variable is *not* merely a statistical aid, but rather a *value judgement* about the nature of the latent variable and the noise in the observations. We are *defining* the noise to be that part of the variability which is uncorrelated with the instrumental variable, and the signal to be that part which is correlated.

In our example above, a good instrumental variable $\mathbf{y}_i$ is a different (noisy) view of the same underlying low-dimensional latent variable: $\mathbf{y}_i = \mathbf{N}\mathbf{z}_i + \boldsymbol{\zeta}_i$ for some full-column-rank linear transformation $\mathbf{N}$ and i.i.d. zero-mean noise term $\boldsymbol{\zeta}_i$. The expectation of the empirical cross covariance $\hat{\boldsymbol{\Sigma}}_{XY} = \frac{1}{n}\mathbf{X}\mathbf{Y}^\mathsf{T}$ is then $\mathbf{M}\operatorname{Cov}(\mathbf{z}_i)\mathbf{N}^\mathsf{T}$: the noise terms, being independent and zero-mean, cancel out. (And the variance of each element of $\hat{\boldsymbol{\Sigma}}_{XY}$ goes to 0 as $n \to \infty$.)

Now, we can identify the embedding by computing the truncated singular value decomposition (SVD) of the covariance: $\langle \mathbf{U}, \mathbf{D}, \mathbf{V} \rangle = \operatorname{SVD}(\hat{\boldsymbol{\Sigma}}_{XY}, k)$. If we set $k$ to be the true dimension of $\mathbf{z}$, then as $n \to \infty$, $\mathbf{U}$ will converge to an orthonormal basis for the range of $\mathbf{M}$, and $\mathbf{V}$ will converge to an orthonormal basis for the range of $\mathbf{N}$. The corresponding embeddings are then given by $\mathbf{U}^\mathsf{T}\mathbf{X}$ and $\mathbf{V}^\mathsf{T}\mathbf{Y}$.[2] Interestingly, we can equally well view $\mathbf{x}_i$ as an instrumental variable for $\mathbf{y}_i$: we simultaneously find consistent embeddings of both $\mathbf{x}_i$ and $\mathbf{y}_i$, using each to unbias the other.

### 3.2. Bias in Nonlinear Models

We now extend the analysis of Section 3.1 to *nonlinear* models. We assume noisy observations $\mathbf{x}_i = f(\mathbf{z}_i) + \boldsymbol{\epsilon}_i$, where $\mathbf{z}_i$ is the desired low-dimensional latent variable, $\boldsymbol{\epsilon}_i$ is an i.i.d. noise term, and $f$ is a smooth function with smooth inverse (so that $f(\mathbf{z}_i)$ lies on a manifold). Our goal is to recover $f$ and $\mathbf{z}_i$ up to identifiability.

Kernel PCA (Sec. 2.1) is a common approach to this problem. In the (restrictive) realizable case, kernel PCA gets the right answer: that is, suppose that $\mathbf{z}_i$ has dimension $k$, that $\boldsymbol{\epsilon}_i$ has zero variance, and that we have at least $k$ independent samples. And, suppose that $\phi(f(\mathbf{z}))$ is a linear function of $\mathbf{z}$. Then, the Gram matrix or the covariance "matrix" will have rank $k$, and we can reconstruct a basis for the range of $\phi \circ$ $f$ from the top $k$ eigenvectors of the Gram matrix. (Similarly, if $\phi \circ f$ is near linear and the variance of $\boldsymbol{\epsilon}_i$ is small, we can expect kernel PCA to work well, if not perfectly.)

---

[2]Several spectral decompositions of cross-covariance matrices can be viewed as special cases of two-subspace PCA that involve transforming the variables $\mathbf{x}_i$ and $\mathbf{y}_i$ before applying a singular value decomposition. Two popular examples are reduced-rank regression (Reinsel & Velu, 1998) and canonical correlation analysis (Hotelling, 1935).

However, just as PCA recovers a biased answer when the variance of $\boldsymbol{\epsilon}_i$ is nonzero, kernel PCA will also recover a biased answer under noise, even in the limit of infinite data. The bias of kernel PCA follows immediately from the example at the beginning of Section 3: if we use a linear kernel, kernel PCA will simply reproduce the bias of ordinary PCA.

#### 3.2.1. INSTRUMENTAL VARIABLES IN HILBERT SPACE

By analogy to two-subspace PCA, a natural generalization of kernel PCA is *two-subspace kernel PCA*, which we can accomplish via a kernelized SVD of a *cross-covariance* operator in Hilbert space. Given a joint distribution $\mathbb{P}[X, Y]$ over two variables $X$ on $\mathcal{X}$ and $Y$ on $\mathcal{Y}$, with feature maps $\phi$ and $\upsilon$ (corresponding to kernels $K_\mathbf{x}$ and $K_\mathbf{y}$), the cross-covariance operator $\boldsymbol{\Sigma}_{XY}$ is $\mathbb{E}[\phi(\mathbf{x}) \otimes \upsilon(\mathbf{y})]$. The cross-covariance operator reduces to an ordinary cross-covariance matrix in the finite-dimensional case; in the infinite-dimensional case, it can be viewed as a *kernel mean map* descriptor (Smola et al., 2007) for the joint distribution $\mathbb{P}[X, Y]$. The concept of a cross-covariance operator allows us to extend the methods of instrumental variables to infinite dimensional RKHSs. In our example above, a good instrumental variable $\mathbf{y}_i$ is a different (noisy) view of the same underlying latent variable: $\mathbf{y}_i = g(\mathbf{z}_i) + \boldsymbol{\zeta}_i$ for some smoothly invertible function $g$ and i.i.d. zero-mean noise term $\boldsymbol{\zeta}_i$.

#### 3.2.2. TWO-SUBSPACE PCA IN RKHSS

We proceed now to derive the kernel SVD for a cross-covariance operator.[3] Conceptually, our inputs are "matrices" $\boldsymbol{\Phi}$ and $\boldsymbol{\Upsilon}$ whose columns are respectively $\phi(\mathbf{x}_i)$ and $\upsilon(\mathbf{y}_i)$. The centered empirical covariance operator is then $\hat{\boldsymbol{\Sigma}}_{XY} = \frac{1}{n}(\boldsymbol{\Phi}\mathbf{H})(\boldsymbol{\Upsilon}\mathbf{H})^\mathsf{T}$. The goal of the kernel SVD is then to factor $\hat{\boldsymbol{\Sigma}}_{XY}$ so that we can recover the desired bases for $\phi(\mathbf{x}_i)$ and $\upsilon(\mathbf{y}_i)$. Unfortunately, this conceptual algorithm is impractical, since $\hat{\boldsymbol{\Sigma}}_{XY}$ can be high- or infinite-dimensional. So, instead, we perform an SVD on the covariance operator in Hilbert space via a trick analogous to kernel PCA.

To understand SVD in general Hilbert spaces, we start by looking at a Gram matrix formulation of finite dimensional SVD. Recall that the singular values of $\hat{\boldsymbol{\Sigma}}_{XY} = \frac{1}{n}(\mathbf{X}\mathbf{H})(\mathbf{Y}\mathbf{H})^\mathsf{T}$ are the square roots of the eigenvalues of $\hat{\boldsymbol{\Sigma}}_{XY}\hat{\boldsymbol{\Sigma}}_{YX}$ (where $\hat{\boldsymbol{\Sigma}}_{YX} = \hat{\boldsymbol{\Sigma}}_{XY}^\mathsf{T}$), and the left singular vectors are defined to be the corresponding eigenvectors. We can find identical eigenvectors and eigenvalues using only centered Gram matrices $\mathbf{C}_X = \frac{1}{n}(\mathbf{XH})^\mathsf{T}(\mathbf{XH})$ and $\mathbf{C}_Y = \frac{1}{n}(\mathbf{YH})^\mathsf{T}(\mathbf{YH})$. Let $\mathbf{v}_i$ be a right eigenvector of $\mathbf{C}_Y\mathbf{C}_X$, so that $\mathbf{C}_Y\mathbf{C}_X\mathbf{v}_i = \lambda_i\mathbf{v}_i$. Premultiplying by $(\mathbf{XH})$ yields

$$\frac{1}{n^2}(\mathbf{XH})(\mathbf{YH})^\mathsf{T}(\mathbf{YH})(\mathbf{XH})^\mathsf{T}(\mathbf{XH})\mathbf{v}_i = \lambda_i(\mathbf{XH})\mathbf{v}_i$$

and regrouping terms gives us $\hat{\boldsymbol{\Sigma}}_{XY}\hat{\boldsymbol{\Sigma}}_{YX}\mathbf{w}_i = \lambda_i\mathbf{w}_i$ where $\mathbf{w}_i = (\mathbf{XH})\mathbf{v}_i$. So, $\lambda_i$ is an eigenvalue of $\hat{\boldsymbol{\Sigma}}_{XY}\hat{\boldsymbol{\Sigma}}_{YX}$, $\sqrt{\lambda_i}$ is a singular value of $\hat{\boldsymbol{\Sigma}}_{XY}$, and $(\mathbf{XH})\mathbf{v}_i\lambda_i^{-1/2}$ is the corresponding unit length left singular vector. An analogous argument shows that, if $\mathbf{w}'_i$ is a unit-length right singular vector of $\hat{\boldsymbol{\Sigma}}_{XY}$, then $\mathbf{w}'_i = (\mathbf{YH})\mathbf{v}'_i\lambda_i^{-1/2}$, where $\mathbf{v}'_i$ is a unit-length left eigenvector of $\mathbf{C}_Y\mathbf{C}_X$.

---

[3]The kernel SVD algorithm previously appeared as an intermediate step in (Song et al., 2010; Fukumizu et al., 2005); here we give a more complete description.

This machinery allows us to solve the two-subspace kernel PCA problem by computing the singular values of the empirical covariance operator $\hat{\boldsymbol{\Sigma}}_{XY}$. We define $\mathbf{G}_X$ and $\mathbf{G}_Y$ to be the Gram matrices whose elements are $K_\mathbf{x}(\mathbf{x}_i, \mathbf{x}_j)$ and $K_\mathbf{y}(\mathbf{y}_i, \mathbf{y}_j)$ respectively, and then compute the eigendecomposition of $\mathbf{C}_Y\mathbf{C}_X = (\mathbf{HG}_Y\mathbf{H})(\mathbf{HG}_X\mathbf{H})$. This method avoids any computations in infinite-dimensional spaces; and, it gives us compact representations of the left and right singular vectors.

Under appropriate assumptions, we can show that the SVD of the empirical cross-covariance operator $\hat{\boldsymbol{\Sigma}}_{XY} = \frac{1}{n}\boldsymbol{\Phi}\mathbf{H}\boldsymbol{\Upsilon}^\mathsf{T}$ converges to the desired value. Suppose that $\mathbb{E}[\phi(\mathbf{x}_i) \mid \mathbf{z}_i]$ is a linear function of $\mathbf{z}_i$, and similarly, that $\mathbb{E}[\upsilon(\mathbf{y}_i) \mid \mathbf{z}_i]$ is a linear function of $\mathbf{z}_i$.[4] The noise terms $\phi(\mathbf{x}_i) - \mathbb{E}[\phi(\mathbf{x}_i) \mid \mathbf{z}_i]$ and $\upsilon(\mathbf{y}_i) - \mathbb{E}[\upsilon(\mathbf{y}_i) \mid \mathbf{z}_i]$ are by definition zero-mean; and they are independent of each other, since the first depends only on $\boldsymbol{\epsilon}_i$ and the second only on $\boldsymbol{\zeta}_i$. So, the noise terms cancel out, and the expectation of $\hat{\boldsymbol{\Sigma}}_{XY}$ is the true covariance $\boldsymbol{\Sigma}_{XY}$. If we additionally assume that the noise terms have finite variance, the product-RKHS norm of the error $\hat{\boldsymbol{\Sigma}}_{XY} - \boldsymbol{\Sigma}_{XY}$ vanishes as $n \to \infty$.

The remainder of the proof follows from the proof of Theorem 1 in (Song et al., 2010) (the convergence of the empirical estimator of the kernel covariance oper-ator). In particular, the top $k$ left singular vectors of $\hat{\boldsymbol{\Sigma}}_{XY}$ converge to a basis for the range of $\mathbb{E}[\phi(\mathbf{x}_i) \mid \mathbf{z}_i]$ (considered as a function of $\mathbf{z}_i$); similarly, the top right singular vectors of $\hat{\boldsymbol{\Sigma}}_{XY}$ converge to a basis for the range of $\mathbb{E}[\upsilon(\mathbf{y}_i) \mid \mathbf{z}_i]$.

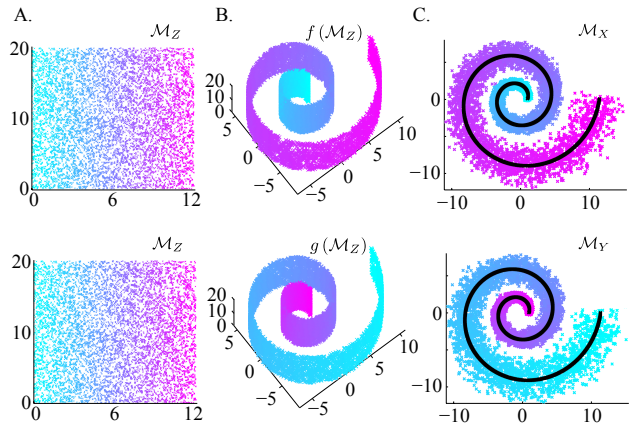

*Figure 1.* The Noisy Swiss Rolls. We are given two sets of 3-d observations residing on two different manifolds $\mathcal{M}_X$ and $\mathcal{M}_Y$. The latent *signal* $\mathcal{M}_Z$ is 2-d, but $\mathcal{M}_X$ and $\mathcal{M}_Y$ are each corrupted by 3-d *noise*. (A) 5000 data points sampled from $\mathcal{M}_Z$. (B) The functions $f(\mathcal{M}_Z)$ and $g(\mathcal{M}_Z)$ "roll" the manifold in 3-dimensional space, two different ways, generating two different manifolds in observation space. (C) Each set of observations is then perturbed by 3-d noise (showing 2 dimensions only), resulting in 3-d manifolds $\mathcal{M}_X$ and $\mathcal{M}_Y$. The black lines indicate the location of the submanifolds from (B).

## 4. Two-Manifold Problems

Now that we have extended the instrumental variable idea to RKHSs, we can also expand the scope of manifold learning to *two-manifold problems*, where we want to simultaneously learn two manifolds for two covarying lists of observations, each corrupted by uncorrelated noise.[5] The idea is simple: we view manifold learners as constructing Gram matrices as in Sec. 2.2, then apply the RKHS instrumental variable idea of Sec. 3. As we will see, this procedure allows us to regain good performance when observations are noisy.

Suppose we are given two set of observations residing on (or near) two different manifolds: $\mathbf{x}_1, \ldots, \mathbf{x}_n \in \mathbb{R}^{d_1}$ on $\mathcal{M}_X$ and $\mathbf{y}_1, \ldots, \mathbf{y}_n \in \mathbb{R}^{d_2}$ on $\mathcal{M}_Y$. Further suppose that both $\mathbf{x}_i$ and $\mathbf{y}_i$ are *noisy* functions of a latent variable $\mathbf{z}_i$, itself residing on a latent $k$-dimensional manifold $\mathcal{M}_Z$: $\mathbf{x}_i = f(\mathbf{z}_i) + \boldsymbol{\epsilon}_i$ and $\mathbf{y}_i = g(\mathbf{z}_i) + \boldsymbol{\zeta}_i$. We assume that the functions $f$ and $g$ are smooth, so that $f(\mathbf{z}_i)$ and $g(\mathbf{z}_i)$ trace out submanifolds $f(\mathcal{M}_Z) \subseteq \mathcal{M}_X$ and $g(\mathcal{M}_Z) \subseteq \mathcal{M}_Y$. We further assume that the noise terms $\boldsymbol{\epsilon}_i$ and $\boldsymbol{\zeta}_i$ move $\mathbf{x}_i$ and $\mathbf{y}_i$ *within* their respective manifolds $\mathcal{M}_X$ and $\mathcal{M}_Y$: this assumption is without loss of generality, since we can can always increase the dimension of the manifolds $\mathcal{M}_X$ and $\mathcal{M}_Y$ to allow an arbitrary noise term. See Figure 1 for an example.

---

[4] The assumption of linearity is restrictive, but appears necessary: in order to learn a representation of a manifold using factorization-based methods, we need to pick a kernel which flattens out the manifold into a subspace. This is why kernel eigenmap methods are generally more successful than plain kernel PCA: by learning an appropriate kernel, they are able to adapt their nonlinearity to the shape of the target manifold.

[5] The uncorrelated noise assumption is extremely mild: if some latent variable causes correlated changes in our measurements on the two manifolds, then we are making the definition that it is part of the desired signal to be recovered. No other definition seems reasonable: if there is no difference in statistical behavior between signal and noise, then it is impossible to use a statistical method to separate signal from noise.

**Algorithm 1** Instrumental Eigenmaps
---
**In**: $n$ i.i.d. pairs of observations $\{\mathbf{x}_i, \mathbf{y}_i\}_{i=1}^n$
**Out**: embeddings $\mathbf{E}_X$ and $\mathbf{E}_Y$
 1: Compute Gram matrices: $\mathbf{G}_X$ and $\mathbf{G}_Y$ from $\mathbf{x}_{1:n}$ and $\mathbf{y}_{1:n}$ respectively, using for example LE.
 2: Compute centered Gram matrices:
    $\mathbf{C}_X = \mathbf{H}\mathbf{G}_X\mathbf{H}$ and $\mathbf{C}_Y = \mathbf{H}\mathbf{G}_Y\mathbf{H}$
 3: Perform a singular value decomposition and truncate the top $k$ singular values:
    $\langle \mathbf{U}, \mathbf{\Lambda}, \mathbf{V}^\mathsf{T} \rangle = \text{SVD}(\mathbf{C}_X\mathbf{C}_Y, k)$
 4: Find the embeddings from the singular values:
    $\mathbf{E}_X = \mathbf{U}_{1:k}\mathbf{\Lambda}_{1:k}^{1/2}$ and
    $\mathbf{E}_Y = \mathbf{V}_{1:k}\mathbf{\Lambda}_{1:k}^{1/2}$

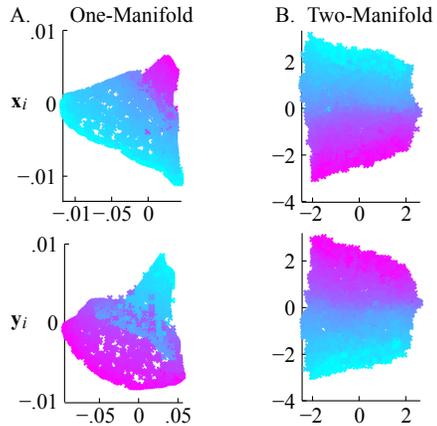

*Figure 2.* Solving the Noisy Swiss Roll two-manifold problem (see Fig. 1 for setup). Top graphs show embeddings of $\mathbf{x}_i$, bottom graphs show embeddings of $\mathbf{y}_i$. (A) 2-dimensional embeddings found by normalized LE. The best results were obtained by setting the number of nearest neighbors to 5. Due to the large amounts of noise, the separately learned embeddings do not accurately reflect the latent 2-dimensional manifold. (B) The embeddings learned from the left and right eigenvectors of $\mathbf{C}_X\mathbf{C}_Y$ closely match the original data sampled from the true manifold. By solving a two-manifold problem, noise disappears in expectation and the latent manifold is recovered.

If the variance of the noise terms $\boldsymbol{\epsilon}_i$ and $\boldsymbol{\zeta}_i$ is too high, or if $\mathcal{M}_X$ and $\mathcal{M}_Y$ are higher-dimensional than the latent $\mathcal{M}_Z$ manifold (i.e., if the noise terms move $\mathbf{x}_i$ and $\mathbf{y}_i$ away from $f(\mathcal{M}_Z)$ and $g(\mathcal{M}_Z)$), then it may be difficult to reconstruct $f(\mathcal{M}_Z)$ or $g(\mathcal{M}_Z)$ separately from $\mathbf{x}_i$ or $\mathbf{y}_i$. Our goal, therefore, is to use $\mathbf{x}_i$ and $\mathbf{y}_i$ together to reconstruct both manifolds simultaneously: the extra information from the correspondence between $\mathbf{x}_i$ and $\mathbf{y}_i$ will make up for noise, allowing success in the two-manifold problem where the individual one-manifold problems are intractable.

Given samples of $n$ i.i.d. pairs $\{\mathbf{x}_i, \mathbf{y}_i\}_{i=1}^n$ from two manifolds, we propose a two-step spectral learning algorithm for two-manifold problems: first, use either a given kernel or an ordinary one-manifold algorithm such as LE or LLE to compute centered Gram matrices $\mathbf{C}_X$ and $\mathbf{C}_Y$ from $\mathbf{x}_i$ and $\mathbf{y}_i$ separately. Second, use kernel SVD to recover the embedding of points in $\mathcal{M}_Z$. The procedure, called instrumental eigenmaps, is summarized in Algorithm 1.

As shown in Figure 2, computing eigenvalues of $\mathbf{C}_X\mathbf{C}_Y$ instead of $\mathbf{C}_X$ or $\mathbf{C}_Y$ alone alters the eigensystem: it promotes directions within each individual learned manifold that are useful for predicting coordinates on the other learned manifold, and demotes directions that are not useful. This effect strengthens our ability to recover relevant dimensions in the face of noise.

## 5. Two-Manifold Detailed Example: Nonlinear System Identification

A fundamental problem in machine learning and robotics is dynamical system identification. This task requires two related subtasks: 1) learning a low dimensional state space, which is often known to lie on a *manifold*; and 2) learning the system dynamics.

We propose tackling this problem by combining two-manifold methods (for task 1) with spectral learning algorithms for nonlinear dynamical systems (for task 2) (Song et al., 2010; Siddiqi et al., 2010; Boots et al., 2011; Boots & Gordon, 2010; Hsu et al., 2009; Boots et al., 2010). Here, we focus on a specific example: we show how to combine HSE-HMMs (Song et al., 2010), a powerful nonparametric approach to system identification, with manifold learning. We demonstrate that the resulting *manifold* HSE-HMM can outperform standard HSE-HMMs (and many other well-known methods for learning dynamical systems): the manifold HSE-HMM accurately discovers a curved low-dimensional manifold which contains the state space, while other methods discover only a (potentially much higher-dimensional) subspace which contains this manifold.

### 5.1. Hilbert Space Embeddings of HMMs
The key idea behind spectral learning of dynamical systems is that a good latent state is one that lets us predict the future. HSE-HMMs implement this idea by finding a low-dimensional embedding of the conditional probability distribution of sequences of future observations, and using the embedding coordinates as state. Song et al. (2010) suggest finding this low-dimensional state space as a subspace of an infinite dimensional RKHS.

Intuitively, we might think that we could find the best state space by performing PCA or kernel PCA of sequences of future observations. That is, we would sample $n$ sequences of future observations $\mathbf{x}_1, \ldots, \mathbf{x}_n \in \mathbb{R}^{d_1}$ from a dynamical system. We would then construct a Gram matrix $\mathbf{G}_X$, whose $(i,j)$ element is $K_\mathbf{x}(\mathbf{x}_i, \mathbf{x}_j)$. Finally, we would find the eigendecomposition of the centered Gram matrix $\mathbf{C}_X = \mathbf{H}\mathbf{G}_X\mathbf{H}$ as in Section 2.1. The resulting embedding coordinates would be tuned to predict future observations well, and so could be viewed as a good state space. However, the state space found by kernel PCA is *biased*: it typically includes *noise*, information that cannot be predicted from past observations. We would like instead to find a low dimensional state space that is *uncorrelated* with the noise in the future observations.

So, in addition to sampling sequences of future observations, we sample corresponding sequences of *past* observations $\mathbf{y}_1, \ldots, \mathbf{y}_n \in \mathbb{R}^{d_2}$: sequence $\mathbf{y}_i$ ends at time $t_i - 1$. We view features of the past as *instrumental variables* to unbias the future. We therefore construct a Gram matrix $\mathbf{G}_Y$, whose $(i,j)$ element is $K_\mathbf{y}(\mathbf{y}_i, \mathbf{y}_j)$. From $\mathbf{G}_Y$ we construct the centered Gram matrix $\mathbf{C}_Y = \mathbf{H}\mathbf{G}_Y\mathbf{H}$. Finally, we identify the state space using a kernel SVD as in Section 3.2.2: $\langle \mathbf{U}, \mathbf{\Lambda}, \mathbf{V}^\mathsf{T} \rangle = \text{SVD}(\mathbf{C}_X\mathbf{C}_Y, k)$. The left singular "vectors" (reconstructed from $\mathbf{U}$ as in Section 3.2.2) now identify a subspace in which the system evolves. From this subspace, we can proceed to identify the parameters of the system as in Song et al. (2010).

### 5.2. Manifold HSE-HMMs

In contrast with ordinary HSE-HMMs, we are interested in modeling a dynamical system whose state space lies on a low-dimensional manifold, even if this manifold is curved to occupy a higher-dimensional subspace (an example is given in Section 5.3, below). We want to use this additional knowledge to constrain the learning algorithm and produce a more accurate model for a given amount of training data. To do so, we replace the kernel SVD by a two-manifold method. That is, we learn centered Gram matrices $\mathbf{C}_X$ and $\mathbf{C}_Y$ for the future and past observations, using a manifold method like LE or LLE (see Section 2.2). Then we apply a SVD to $\mathbf{C}_X\mathbf{C}_Y$ in order to recover the latent state space.

### 5.3. Slotcar: A Real-World Dynamical System

To evaluate two-manifold HSE-HMMs we look at the problem of tracking and predicting the position of a slotcar with attached inertial measurement unit (IMU) racing around a track. Figure 3(A) shows setup.

We collected 3000 successive observations of 3D acceleration and angular velocity at 10 Hz while the slot car circled the track controlled by a constant policy (with varying speeds). The goal was to learn a dynamical model of the noisy IMU data, and, after filtering, to predict current and future 2-dimensional locations. We used the first 2000 data points as training data, and held out the last 500 data points for testing the learned models. We trained four models, and evaluated these models based on prediction accuracy, and, where appropriate, the learned latent state.

First, we trained a 20-dimensional embedded HMM with the spectral algorithm of Song et al. (2010), using sequences of 150 consecutive observations and Gaussian RBF kernels. Second, we trained a similar 20-dimensional embedded HMM with normalized LE kernels. (Normalized LE differs from LE by utilizing the normalized graph Laplacian instead of the standard graph Laplacian.) The number of nearest neighbors was selected to be 50, and the other parameters were set to be identical to the first model. (So, the only difference is that the first model performs a kernel SVD, while the second model solves a two-manifold problem.) Third, we trained a 20-dimensional Kalman filter using the N4SID algorithm (Van Overschee & De Moor, 1996) with Hankel matrices of 150 time steps; and finally, we learned a 20-state HMM (with 400 levels of discretization for observations) via the EM algorithm.

We compared the learned state spaces of the first three models. These models differ mainly in their kernel: Gaussian RBF, learned manifold from normalized LE, or linear. As a test, we tried to reconstruct the 2-dimensional locations of the car (collected from an overhead camera, and not used in learning the dynamical systems) from each of the three latent state spaces: the more accurate the learned state space, the better we expect to be able to reconstruct the locations. Results are shown in Figure 3(B).

Finally we examined the prediction accuracy of each model. We performed filtering for different extents $t_1 = 100, \ldots, 350$, then predicted the car location for a further $t_2$ steps in the future, for $t_2 = 1, \ldots, 100$. The root-mean-squared error of this prediction in the 2-dimensional location space is plotted in Figure 3(C). The Manifold HMM learned by the method detailed in Section 5.2 consistently yields lower prediction error

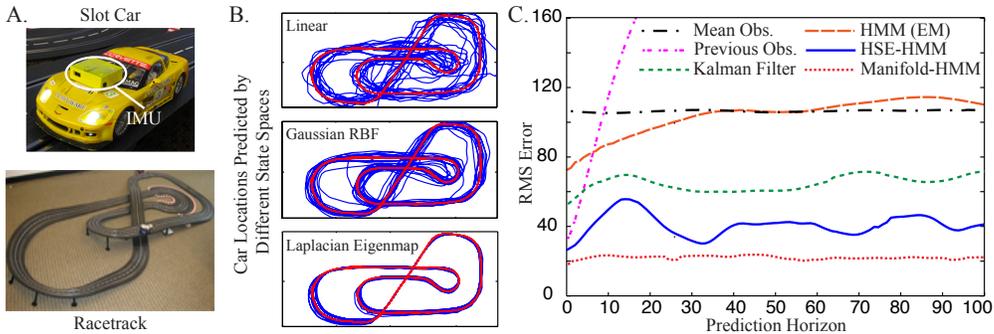

*Figure 3.* The slot car platform. (A) The car and inertial measurement unit (IMU) (top) and racetrack (bottom). (B) A comparison of training data embedded into the state space of three different learned models. Red line indicates true 2-d position of the car over time, blue lines indicate the prediction from state space. The top graph shows the Kalman filter state space (linear kernel), the middle graph shows the HSE-HMM state space (RBF kernel), and the bottom graph shows the manifold HSE-HMM state space (LE kernel). The LE kernel finds the best representation of the true manifold. (C) RMS error for prediction (averaged over 251 time steps) with different estimated models. The HSE-HMM significantly outperforms the other learned models by taking advantage of the fact that the data we want to predict lies on a manifold.

for the duration of the prediction horizon.

## 6. Related Work

While preparing this manuscript, we learned of the simultaneous and independent work of Mahadevan et al. (2011). That paper defines one particular two-manifold algorithm, maximum covariance unfolding (MCU). We believe the current paper will help to elucidate *why* two-manifold methods like MCU work well.

A similar problem to the two-manifold problem is *manifold alignment* (Ham et al., 2005; Wang & Mahadevan, 2009), which builds connections between two or more data sets by aligning their underlying manifolds. Our aim is different: we assume paired data, where manifold alignments do not; and, we focus on learning algorithms that *simultaneously* discover manifold structure and connections between manifolds (as provided by, e.g., a top-level learning problem defined between two manifolds).

Interconnected dimensionality reduction has been explored before in sufficient dimension reduction (SDR) (Li, 1991; Cook & Yin, 2001; Fukumizu et al., 2004). In SDR, the goal is to find a linear subspace of covariates $\mathbf{x}_i$ that makes response vectors $\mathbf{y}_i$ conditionally independent of the $\mathbf{x}_i$s. The formulation is in terms of conditional independence. Unfortunately, the solution to SDR problems usually requires a difficult nonlinear non-convex optimization. A related method is manifold kernel dimension reduction (Nilsson et al., 2007), which finds an embedding of covariates $\mathbf{x}_i$ using a kernel eigenmap method, and then attempts to find a linear transformation of some of the dimensions of the embedded points to predict response variables $\mathbf{y}_i$. The response variables are constrained to be linear in the manifold, so the problem is quite different from a two-manifold problem.

## 7. Conclusion

In this paper we study *two-manifold problems*, where two sets of corresponding data points, generated from a single latent manifold and corrupted by noise, lie on or near two different higher dimensional manifolds. We design algorithms by relating two-manifold problems to cross-covariance operators in RKHSs, and show that these algorithms result in a significant improvement over standard manifold learning approaches in the presence of noise. This is an appealing result: manifold learning algorithms typically assume that observations are (close to) noiseless, an assumption that is rarely satisfied in practice.

Furthermore, we demonstrate the utility of two-manifold problems by extending a recent dynamical system identification algorithm to learn a system with a state space that lies on a manifold. The resulting algorithm learns a model that outperforms the current state of the art in predictive accuracy. To our knowledge this is the first combination of system identification and manifold learning that accurately identifies a latent time series manifold *and* is competitive with the best system identification algorithms at learning accurate predictive models.

### Acknowledgements
Byron Boots and Geoffrey J. Gordon were supported by ONR MURI grant number N00014-09-1-1052. Byron Boots was supported by the NSF under grant number EEEC-0540865.

## References

Belkin, Mikhail and Niyogi, Partha. Laplacian eigenmaps for dimensionality reduction and data representation. *Neural Computation*, 15:1373–1396, 2002.

Boots, Byron and Gordon, Geoff. Predictive state temporal difference learning. In Lafferty, J., Williams, C. K. I., Shawe-Taylor, J., Zemel, R.S., and Culotta, A. (eds.), *Advances in Neural Information Processing Systems 23*, pp. 271–279. 2010.

Boots, Byron, Siddiqi, Sajid M., and Gordon, Geoffrey J. Closing the learning-planning loop with predictive state representations. In *Proceedings of Robotics: Science and Systems VI*, 2010.

Boots, Byron, Siddiqi, Sajid, and Gordon, Geoffrey. An online spectral learning algorithm for partially observable nonlinear dynamical systems. In *Proceedings of the 25th National Conference on Artificial Intelligence (AAAI-2011)*, 2011.

Bradtke, Steven J. and Barto, Andrew G. Linear least-squares algorithms for temporal difference learning. In *Machine Learning*, pp. 22–33, 1996.

Chen, Guisheng, Yin, Junsong, and Li, Deyi. Neighborhood smoothing embedding for noisy manifold learning. In *GrC*, pp. 136–141, 2008.

Cook, R. Dennis and Yin, Xiangrong. Theory and methods: Special invited paper: Dimension reduction and visualization in discriminant analysis (with discussion). *Australian and New Zealand Journal of Statistics*, 43(2): 147–199, 2001.

Fukumizu, K., Bach, F., and Gretton, A. Consistency of kernel canonical correlation analysis. Technical Report 942, Institute of Statistical Mathematics, Tokyo, Japan, 2005.

Fukumizu, Kenji, Bach, Francis R., Jordan, Michael I., and Williams, Chris. Dimensionality reduction for supervised learning with reproducing kernel Hilbert spaces. *Journal of Machine Learning Research*, 5:2004, 2004.

Ham, Jihun, Lee, Daniel D., Mika, Sebastian, and Schlkopf, Bernhard. A kernel view of the dimensionality reduction of manifolds, 2003.

Ham, Jihun, Lee, Daniel, and Saul, Lawrence. Semisupervised alignment of manifolds. In Cowell, Robert G. and Ghahramani, Zoubin (eds.), *10th International Workshop on Artificial Intelligence and Statistics*, pp. 120–127, 2005.

Hotelling, Harold. The most predictable criterion. *Journal of Educational Psychology*, 26:139–142, 1935.

Hsu, Daniel, Kakade, Sham, and Zhang, Tong. A spectral algorithm for learning hidden Markov models. In *COLT*, 2009.

Jolliffe, I. T. *Principal component analysis*. Springer, New York, 2002.

Lawrence, Neil D. Spectral dimensionality reduction via maximum entropy. In *Proc. AISTATS*, 2011.

Li, Ker-Chau. Sliced inverse regression for dimension reduction. *Journal of the American Statistical Association*, 86(414):pp. 316–327, 1991.

Mahadevan, Vijay, Wong, Chi Wah, Pereira, Jose Costa, Liu, Tom, Vasconcelos, Nuno, and Saul, Lawrence. Maximum covariance unfolding: Manifold learning for bimodal data. In *Advances in Neural Information Processing Systems 24*, 2011.

Nilsson, Jens, Sha, Fei, and Jordan, Michael I. Regression on manifolds using kernel dimension reduction. In *ICML*, pp. 697–704, 2007.

Pearl, Judea. *Causality: models, reasoning, and inference*. Cambridge University Press, 2000.

Reinsel, Gregory C. and Velu, Rajabather Palani. *Multivariate Reduced-rank Regression: Theory and Applications*. Springer, 1998.

Roweis, Sam T. and Saul, Lawrence K. Nonlinear dimensionality reduction by locally linear embedding. *Science*, 290(5500):2323–2326, December 2000.

Schölkopf, Bernhard, Smola, Alex J., and Müller, Klaus-Robert. Nonlinear component analysis as a kernel eigenvalue problem. *Neural Computation*, 10(5):1299–1319, 1998.

Siddiqi, Sajid, Boots, Byron, and Gordon, Geoffrey J. Reduced-rank hidden Markov models. In *Proceedings of the Thirteenth International Conference on Artificial Intelligence and Statistics (AISTATS-2010)*, 2010.

Smola, A.J., Gretton, A., Song, L., and Schölkopf, B. A Hilbert space embedding for distributions. In Takimoto, E. (ed.), *Algorithmic Learning Theory*, Lecture Notes on Computer Science. Springer, 2007.

Song, L., Boots, B., Siddiqi, S. M., Gordon, G. J., and Smola, A. J. Hilbert space embeddings of hidden Markov models. In *Proc. 27th Intl. Conf. on Machine Learning (ICML)*, 2010.

Tenenbaum, Joshua B., Silva, Vin De, and Langford, John. A global geometric framework for nonlinear dimensionality reduction. *Science*, 290:2319–2323, 2000. doi: 10.1126/science.290.5500.2319.

Van Overschee, P. and De Moor, B. *Subspace Identification for Linear Systems: Theory, Implementation, Applications*. Kluwer, 1996.

Wang, Chang and Mahadevan, Sridhar. A general framework for manifold alignment. In *Proc. AAAI*, 2009.

Weinberger, Kilian Q., Sha, Fei, and Saul, Lawrence K. Learning a kernel matrix for nonlinear dimensionality reduction. In *In Proceedings of the 21st International Conference on Machine Learning*, pp. 839–846. ACM Press, 2004.

Zhan, Yubin and Yin, Jianping. Robust local tangent space alignment. In *ICONIP (1)*, pp. 293–301, 2009.

Zhan, Yubin and Yin, Jianping. Robust local tangent space alignment via iterative weighted PCA. *Neurocomputing*, 74(11):1985–1993, 2011.